\documentclass[10pt, conference, compsocconf]{IEEEtran}
\ifCLASSINFOpdf
  \usepackage[pdftex]{graphicx}
\else
\fi
\usepackage[cmex10]{amsmath}
\usepackage{amssymb}

\usepackage[caption=false,font=scriptsize]{subfig}

\usepackage{url}

\newcommand{\eg}{\textit{e.g.~}}
\newcommand{\ie}{\textit{i.e.~}}

\begin{document}
\sloppy
%
\title{Evaluation of Dense 3D Reconstruction from 2D Face Images in the Wild}


\author{Zhen-Hua Feng$^{1}$ ~~Patrik Huber$^1$ ~~Josef Kittler$^1$ ~~Peter JB Hancock$^2$ ~~Xiao-Jun Wu$^3$ \\  ~~Qijun Zhao$^4$ ~~Paul Koppen$^1$ ~~Matthias R\"atsch$^5$\\
$^1$Centre for Vision, Speech and Signal Processing, University of Surrey, Guildford GU2 7XH, UK\\
$^2$Faculty of Natural Sciences, University of Stirling, Stirling FK9 4LA, UK\\
$^3$School of Internet of Things Engineering, Jiangnan University, Wuxi 214122, China\\
$^4$Biometrics Research Lab, College of Computer Science, Sichuan University, Chengdu 610065, China \\
$^5$Image Understanding and Interactive Robotics, Reutlingen University, 72762 Reutlingen, Germany\\
{\tt\small \{z.feng, j.kittler, p.koppen\}@surrey.ac.uk, patrikhuber@gmail.com,}
\\ 
{\tt\small wu\_xiaojun@jiangnan.edu.cn, p.j.b.hancock@stir.ac.uk,  qjzhao@scu.edu.cn}
}


%


\maketitle

\begin{abstract}
This paper investigates the evaluation of dense 3D face reconstruction from a single 2D image in the wild.
To this end, we organise a competition that provides a new benchmark dataset that contains 2000 2D facial images of 135 subjects as well as their 3D ground truth face scans.
In contrast to previous competitions or challenges, the aim of this new benchmark dataset is to evaluate the accuracy of a 3D dense face reconstruction algorithm using \emph{real}, accurate and high-resolution 3D ground truth face scans.
In addition to the dataset, we provide a standard protocol as well as a Python script for the evaluation. 
Last, we report the results obtained by five state-of-the-art 3D face reconstruction systems on the new benchmark dataset.
The competition is organised along with the 2018 13th IEEE Conference on Automatic Face \& Gesture Recognition.
\emph{This is an extended version of the original conference submission with two additional 3D face reconstruction systems evaluated on the benchmark.}
\end{abstract}


%
\IEEEpeerreviewmaketitle

\section{Introduction}
3D face reconstruction from 2D images is a very active topic in many research areas such as computer vision, pattern recognition and computer graphics~\cite{DBLP:conf/siggraph/BlanzV99,feng2015cascaded,kittler20163d,zeng2017examplar,DBLP:conf/icb/LiuHSWZ17,koppen2018gaussian}. 
While the topic has been researched for nearly two decades (with one of the seminal papers being Blanz \& Vetter~\cite{DBLP:conf/siggraph/BlanzV99}), over the last two years, these methods have been growing out of laboratory-applications and have become applicable to in-the-wild images, containing larger pose variations, difficult illumination conditions, facial expressions, or different ethnicity groups and age ranges~\cite{DBLP:conf/cvpr/ZhuLLSL16,DBLP:conf/iccv/BulatT17,liu2017dense}. 
However, it is currently an ongoing challenge to quantitatively evaluate such algorithms: For 2D images captured in the wild, there is usually no 3D ground truth available. And, vice-versa, for 3D data, it is usually captured with a 3D scanner in a laboratory, and no in-the-wild 2D images of the same subject exist.

Thus, researchers often publish qualitative results, alongside some effort of quantitative evaluation, which is often not ideal, for the lack of 3D ground truth. Also, people resort to using a proxy task, for example face recognition. 
For an example of the former, people are using the Florence 2D/3D hybrid face dataset (MICC)~\cite{Bagdanov:2011:FHF:2072572.2072597}, which contains 3D data with 2D videos (\eg \cite{DBLP:journals/cg/HernandezHCM17,DBLP:conf/cvpr/TranHMM17}). However there is no standard evaluation protocol, and very often synthetic renderings of the 3D scans are used for evaluation, which do not contain background (\eg they are rendered on black background) or any natural illumination variations (\eg \cite{DBLP:journals/corr/JacksonBAT17}), for the lack of better data. Other methods (\eg 3DDFA~\cite{DBLP:conf/cvpr/ZhuLLSL16}) compare their results against a `ground truth' which is created by another fitting algorithm, which is itself problematic as these fitting algorithms have not yet been shown to be effective on in-the-wild data, even after manual correction. There have been a limited number of previous competitions which aimed to improve the situation, but they only solved the problem partially. For example the workshop organised by Jeni et al.~\cite{DBLP:conf/eccv/JeniTYSC16} used their own algorithm as `ground truth' (see also Section~\ref{sec:previous_workshops}). Other datasets have been recently proposed like KF-ITW~\cite{DBLP:journals/corr/BoothAPTPZ17} but therein Kinect Fusion is used as 3D ground truth, which does not consist of very high resolution meshes, and also the videos are recorded in rather controlled and similar scenarios (\ie rotating around a chair in a lab).

In this paper, we report the results of a competition on 3D dense face reconstruction of in-the-wild 2D images, evaluated with accurate and high-resolution 3D ground truth, obtained from a 3D structured-light system.
The competition is co-located with a workshop\footnote{\url{https://facer2vm.org/fg2018/}} of the 13th IEEE Conference on Automatic Face \& Gesture Recognition (FG 2018).


\subsection{Outcomes}
\begin{itemize}
    \item The competition provides a benchmark dataset with 2000 2D images of 135 subjects as well as their high-resolution 3D ground truth face scans.
    Alongside the dataset we supply a standard benchmark protocol to be used on the dataset, for future evaluations and comparisons, beyond the competition.
    
    \item An independent, objective evaluation and comparison of state-of-the art 3D face reconstruction algorithms. The plan is to perform two sets of evaluations: One set for single-image reconstruction, and another set where it is allowed to use all images of one particular person to reconstruct the 3D shape, allowing algorithms to leverage information from multiple images.
    Note that, in this paper, we only report results of the single image fitting protocol.
    
\end{itemize}

\subsection{Impact}
This is the first challenge in 3D face reconstruction from single 2D in-the-wild images with \emph{real}, accurate and high-resolution 3D ground truth.

The provided benchmark dataset is publicly available, so that it can become a benchmark and reference point for future evaluations in the community.

The multi-image challenge allows to test algorithms that can work with multiple videos as well, having far-reaching impact, for example also in the face recognition community (\eg for set-to-set matching, and recent 2D face recognition benchmarks such as the IARPA Janus Benchmark face challenge\footnote{\url{https://www.nist.gov/programs-projects/face-challenges}}).

In addition to that, one of the baseline 3D reconstruction algorithms and the \emph{Surrey Face Model} (SFM) is publicly available too~\cite{DBLP:conf/visapp/HuberHTMKCRK16}.

\subsection{Relationship to previous workshops (competitions)} \label{sec:previous_workshops}

The topic of evaluating 3D face reconstruction algorithms on 2D in-the-wild data has gained much traction recently. The \emph{1st Workshop on 3D Face Alignment in the Wild (3DFAW) Challenge}\footnote{\url{http://mhug.disi.unitn.it/workshop/3dfaw/}}~\cite{DBLP:conf/eccv/JeniTYSC16} was held at ECCV 2016. The benchmark consisted of images from Multi-PIE, synthetically rendered images, and some in-the-wild images from the internet. The 3D `ground truth' was generated by an automatic algorithm provided by the organisers.

As part of ICCV 2017, the iBUG group from Imperial College, UK, held a workshop \emph{1st 3D Face Tracking in-the-wild Competition}\footnote{\url{https://ibug.doc.ic.ac.uk/resources/1st-3d-face-tracking-wild-competition/}}. It improved upon the ECCV 2016 challenge in some respects, but the `ground truth' used was still from an automatic fitting algorithm, introducing bias, and resulting in the other algorithms being evaluated against the performance of another algorithm, and not against real 3D ground truth. Also, the evaluation is only done on a set of sparse 2D and 3D landmarks and not over a dense 3D mesh, leaving much room for further improvements on the benchmarking methodology.

The remaining of this paper outlines the data, protocol, evaluation metrics and results of the competition.
The aim of the competition is to evaluate 3D face shape reconstruction performance of participants on true 2D in-the-wild images, with actual 3D ground truth available from 3D face scanners.
The data is released to the public, together with a well-defined protocol, to provide a standard and public benchmark to the 3D face reconstruction community.

\section{Datasets}
In general, the data used for the evaluation of a 3D face reconstruction algorithm should consist of a number of high-resolution 3D face scans, obtained from a 3D face imaging device, like for example the 3dMDface\footnote{\url{http://www.3dmd.com/}} system.
Together with these 3D face ground truth, associated with each subject are multiple 2D images that have been captured in-the-wild, with a variety of appearance variations in pose, expression, illumination and occlusion.
The aim is to measure the accuracy of an algorithm in reconstructing a subject's neutral 3D face mesh from unconstrained 2D images.
To this end, the Stirling ESRC 3D face dataset\footnote{\url{http://pics.stir.ac.uk/}} is used to create the test set and the JNU~\cite{koppen2018gaussian} 3D face dataset is used to form the validation set.
For training, any 2D or 3D face dataset is allowed except for the Stirling ESRC and JNU datasets.
\begin{figure}[!t]
\centering
\subfloat[high quality images]{
 \label{fig:stirling_data_hq}
 \includegraphics[clip,trim=0 0 0 0, width=.98\linewidth]{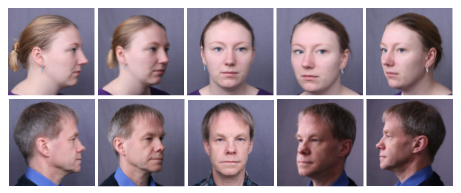}
}

\subfloat[low quality images]{
 \label{fig:stirling_data_lq}
 \includegraphics[clip,trim=0 0 0 0, width=.98\linewidth]{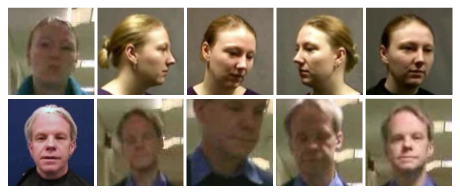}
}
\caption{Some examples of the 2D images in the test set, selected from the Stirling ESRC 3D face dataset.}
\label{fig:stirling_data}
\end{figure}

\subsection{Test set}
The test set is a subset of the Stirling ESRC face database that contains a number of 2D images, video sequences as well as 3D face scans of more than 100 subjects.
The 2D and 3D faces in the Stirling ESRC dataset were captured under 7 different expression variations.
To create the test set, 2000 2D neutral face images, including 656 high-quality and 1344 low-quality images, of 135 subjects were selected from the Stirling ESRC 3D face database.
The high quality images were captured in constrained scenarios with good lighting conditions. The resolution of a high quality face image is higher than 1000$\times$1000.
In contrast, the low quality images or video frames were captured with a variety of image degradation types such as image blur, low resolution, poor lighting, large scale pose rotation etc.
Some examples of the selected 2D face images are shown in Fig.~\ref{fig:stirling_data}.

\subsection{Validation set}
Participants are allowed to use the validation set to fine-tune the hyper-parameters of their 3D face reconstruction systems, if required.
The validation set contains 161 2D in-the-wild images of 10 subjects and their ground-truth 3D face scans. 
The validation set is a part of the JNU 3D face dataset collected by the Jiangnan University using a 3dMDface system.
The full JNU 3D face dataset has the high resolution 3D face scans of 774 Asian subjects.
For more details of the JNU 3D face dataset, please refer to~\cite{koppen2018gaussian}.





\section{Protocol \& evaluation metrics} 
\label{sec:Protocol}
This section details the exact protocol, rules, and evaluation metrics to be used for the competition.

\subsection{Protocol}
The task of the competition is to reconstruct a subject's neutral 3D face shape from a single 2D input image.
Multi-image fitting that allows the use of multiple input images of the same subject for the task of reconstruction of the subject's neutral face is not included in this competition.
However, we leave this potential evaluation protocol as our future work.

For single image reconstruction, an algorithm is expected to take as input a single 2D image and output a neutral 3D face shape of the subject's identity.
An algorithm should be run on each of the images of each subject individually.
In addition, one set of parameters has to be used for all images of all subjects. No fine-tuning is allowed on a per-image or per-subject basis.


\subsection{Evaluation metrics} \label{sec:evaluation_metrics}
\begin{figure}[!t]
\centering
\includegraphics[clip,trim=0 0 0 0,width=0.16\textwidth]{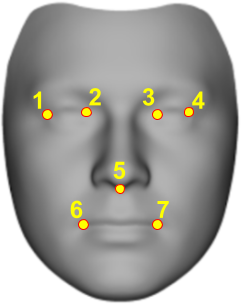}
\hspace{1cm}
\includegraphics[clip,trim=0 0 0 0,width=0.2\textwidth]{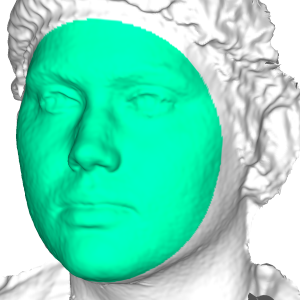}
\caption{\textbf{Left}: The pre-defined seven landmarks used for the rigid alignment of the predicted face mesh with its ground-truth. In order: 1) right eye outer corner, 2) right eye inner corner, 3) left eye inner corner, 4) left eye outer corner, 5) nose bottom, 6) right mouth corner, and 7) left mouth corner.
\textbf{Right}: The area over which face reconstruction is evaluated is defined for each ground-truth 3D scan by a radius around the face centre. This radius is relative to the subject's inter-ocular and eye-nose distance (see Section~\ref{sec:evaluation_metrics} for details).}
\label{fig_face_area}
\end{figure}

Given an input 2D image or a set of 2D images, we use the 3D Root-Mean-Square Error (3D-RMSE) between the reconstructed 3D face shape and the ground truth 3D face scan calculated over an area consisting of the inner face as the evaluation metric.
The area is defined as each vertex in the ground truth scan that is inside the radius from the face centre.
The face centre is computed as the point between the annotated \emph{nose bottom} point and the \emph{nose bridge} (which is computed as the middle of the two annotated eye landmarks): $ \text{face\_centre} = \text{nose\_bottom} + 0.3 \times (\text{nose\_bridge} - \text{nose\_bottom}) $.
The radius is computed as the average of the outer-eye-distance and the distance between nose bridge and nose bottom, times a factor of 1.2:
$ \text{radius} = 1.2 \times (\text{outer\_eye\_dist} + \text{nose\_dist}) / 2 $.
The radius is defined in a relative way because it is desired that the radius covers roughly the same (semantic) area on each scan, and we would like to avoid that e.g. with a very wide face, the evaluated area would cover a smaller part of that particular face.
Typically, the resulting radius is around 80~mm.
The area is depicted for an example scan in Figure~\ref{fig_face_area}.

The following steps are performed to compute the 3D-RMSE between two meshes:
\begin{enumerate}
    \item The predicted and ground truth meshes will be rigidly aligned (by translation, rotation, and scaling). Scaling is compensated for because participants' resulting meshes might be in a different coordinate system, whereas the ground truth scans are in the unit of millimetres. The rigid alignment is based on seven points: both inner and outer eye corners, the nose bottom and the mouth corners (see Fig.~\ref{fig_face_area}).
    The ground truth scans have been annotated with these seven points, whereas participants are expected to specify those on their resulting meshes.
    
    \item For each vertex in the ground truth 3D face scan, the distance is computed to the closest point on the surface of the predicted mesh. 
    These distances are used to compute the 3D-RMSE as well as more specific analysis of a 3D face reconstruction system, \eg the distribution of errors across different face regions.
\end{enumerate}


A Python script is provided\footnote{\url{https://github.com/patrikhuber/fg2018-competition}}, performing the alignment and distance computations.
The output of the script is an ordered list of distances.
\section{Summary of approaches}
Five 3D face reconstruction systems have been evaluated for the competition, including the system submitted by the Biometrics Research Lab at the Sichuan University (SCU-BRL)~\cite{tian_fg2018_workshop}, two systems implemented by  the competition organisers from the University of Surrey and two state-of-the-art deep learning based systems, \ie 3DDFA~\cite{DBLP:conf/cvpr/ZhuLLSL16} and 3DMM-CNN~\cite{DBLP:conf/cvpr/TranHMM17}.
Results were provided in the form of text files with per-vertex errors.
In addition, participants were asked to provide a brief summary of their approach. The descriptions below are based upon these provided summaries.
A thorough comparison will then be presented in Section~\ref{sec:competition_results}.

\subsection{University of Surrey}
The two systems developed by the Centre for Vision, Speech and Signal Processing (CVSSP) from the University of Surrey have three main stages: face detection, facial landmark localisation and 3D face reconstruction.

\subsubsection{Face detection}For face detection, the Multi-Task CNN (MTCNN) face detector was adopted to obtain the bounding box for each input 2D face image~\cite{7553523}. However, the faces of some low resolution images with extreme pose variations were missed by MTCNN. 
For those face images, a bounding box regression approach was used to obtain the face bounding box, as described in~\cite{feng2017face,feng2017dynamic}.

\subsubsection{Facial landmark detection}
\begin{figure}[!t]
\centering
\includegraphics[clip, trim = 27mm 15mm 20mm 10mm, width = .98\linewidth]{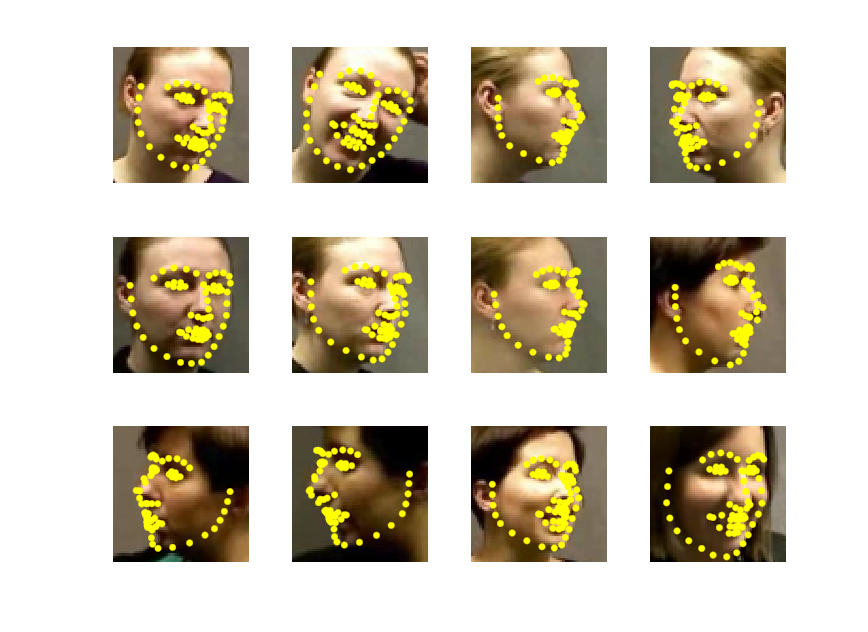}
\caption{Some examples of the detected 68 landmarks by CNN6.}
\label{fig_cnn6_lmks}
\end{figure}

To perform facial landmark localisation, the CNN6 model~\cite{feng2018wing}, a simple CNN-based facial landmark localisation algorithm, was adopted.
The model was trained on multiple in-the-wild face datasets, including the HELEN~\cite{le2012interactive}, LFPW~\cite{belhumeur2013localizing}, ibug~\cite{sagonas2013300} and AFW~\cite{zhu2012face} datasets.
These datasets have 3837 2D face images and each image has 68 facial landmarks annotated by the iBUG group from Imperial College London.
In addition, a subset of the Multi-PIE~\cite{gross2010multi} face dataset was also used for the CNN6 model training.
This Multi-PIE subset has 25,200 2D face images and each face image was manually annotated using 68 facial landmarks~\cite{feng2016unified}.
Some examples from the Stirling low quality subset with the detected 68 facial landmarks using the CNN6 are shown in Fig.~\ref{fig_cnn6_lmks}.

\subsubsection{3D face reconstruction}
Given an input 2D image as well as its 2D facial landmarks, our eos~\cite{DBLP:conf/visapp/HuberHTMKCRK16} and Efficient Stepwise Optimisation (ESO)~\cite{hu2017efficient} fitting algorithms are used to recover the 3D face of the input.
We use the term `MTCNN-CNN6-eos' and `MTCNN-CNN6-ESO' for these two systems.

The eos fitting algorithm reconstructs the 3D face shape based on the landmarks, using a 3D morphable shape and expression model. It consists of shape identity and blendshapes fitting, a scaled orthographic projection camera model, and a dynamic face contour fitting. For this evaluation, the SFM\_3448 shape-only model was used, with the 6 standard Surrey expression blendshapes.
The fitting was run for 5 iterations, fitting all shape coefficients of the model, and with a shape regularisation parameter of $\lambda=30$. 
The eos fitting algorithm is tailored for real-time 3D face fitting applications with a speed of more than 100 fps on a single CPU. It only relies on 2D facial landmarks, and does not use any texture information.

The ESO fitting algorithm is able to reconstruct the 3D shape, texture and lighting from a single 2D face image, using a 3DMM face model.
We use the SFM\_29587 for ESO fitting.
For the shape model, 55 shape eigen-vectors are used.
It should be noted that ESO only uses 2D facial landmarks for shape fitting that does not require any texture information.

\subsection{Sichuan University (SCU-BRL)}
The system developed by the Biometrics Research Lab at the Sichuan University is based on a novel method that is able to reconstruct 3D faces from arbitrary number of 2D images using 2D facial landmarks.
The method is implemented via cascaded regression in shape space. It can effectively exploit complementary information in unconstrained images of varying poses and expressions. It begins with extracting 2D facial landmarks on the images, and then progressively updates the estimated 3D face shape for the input subject via a set of cascaded regressors, which are off-line learned based on a training set of pairing 3D face shapes and unconstrained face images.

\subsubsection{Facial landmark detection}
For 2D facial landmark detection, the state-of-the-art Face Alignment Network (FAN)~\cite{DBLP:conf/iccv/BulatT17} was adopted by in the system developed by SCU-BRL.
For some low quality images, FAN failed, and they manually annotated the 68 landmarks for them. 

\subsubsection{3D face reconstruction}
Given an arbitrary number of unconstrained face images $\{I_{i}\}_{i=1}^{p}, 1\leq p\leq N$ of a subject, the goal is to reconstruct the person-specific frontal and neutral 3D face shape of the subject. 
We represent the 3D face shape by $\mathbf{S} \in \mathbb{R}^{3\times q}$ based on the 3D coordinates of its $q$ vertices, and denote a subset of $\mathbf{S}$ with columns corresponding to $l$ annotated landmarks ($l=68$ in our implementation) as $\mathbf{S}_{L}$. 
The projection of $\mathbf{S}_{L}$ on 2D planes are represented by $\mathbf{U}_{i} \in \mathbb{R}^{2\times l}$. The relationship between 2D facial landmarks $\mathbf{U}_{i}$ and its corresponding 3D landmarks $\mathbf{S}_{L}$ can be described as:
\begin{equation}
\label{eqn::camera_projection}
\mathbf{U}_{i} \approx f_{i}P_{i}R_{i}( \mathbf{S}_{L} + t_{i}),
\end{equation}
where $f_{i}$ is the scale factor, $P_{i}$ is the orthographic projection matrix, $R_{i}$ is the $3 \times 3$ rotation matrix and $t_{i}$ is the translation vector. Here, we employ weak perspective projection $M_{i}$ to approximate the 3D-to-2D mapping. To fully utilize the correlation between the landmarks on all the images, we concatenate them to form a unified 2D facial landmark vector $\mathbf{U}=(\mathbf{U}_{1}, \mathbf{U}_{2}, \cdots, \mathbf{U}_{p}, \mathbf{U}_{p+1}, \cdots, \mathbf{U}_{N})$, where $\mathbf{U}_{i}$ are zero vectors for $(p+1)\leq i \leq N$.

We reconstruct $\mathbf{S}$ from the given `ground truth' landmarks $\mathbf{U}^{*}$ (either manually marked or automatically detected by a standalone method) for the unconstrained image set $\{I_{i}\}_{i=1}^{p}$.  Let $\mathbf{S}^{k-1}$ be the currently reconstructed 3D shape after $k-1$ iterations. The corresponding landmarks $\mathbf{U}^{k-1}$ can be obtained by projecting $\mathbf{S}^{k-1}$ onto the image according to Eqn.~(\ref{eqn::camera_projection}). Then the updated 3D shape $\mathbf{S}^{k}$ can be computed by
\begin{equation}
\label{eqn::shape_increment}
\mathbf{S}^{k} = \mathbf{S}^{k-1}+\mathbf{W}^{k}(\mathbf{U}^{*}-\mathbf{U}^{k-1}),
\end{equation}
where $\mathbf{W}^{k}$ is the regressor in $k^{\texttt{th}}$ iteration.

The $K$ regressors $\{\mathbf{W}^{k}\}_{1}^{K}$ involved in the reconstruction process can be learned via optimising the following objective function over the $m$ training samples (each sample contains up to $N$ annotated 2D images and one ground truth 3D face shape):
\begin{equation}
\label{eqn::objectiveFun}
 \mathop {\arg \min }\limits_{{\mathbf{W}^{k}}}\sum_{j=1}^m\parallel(\mathbf{S}_{j}^{*} - \mathbf{S}^{k-1}_{j}) - \mathbf{W}^{k}(\mathbf{U}_{j}^{*}-\mathbf{U}_{j}^{k-1})\parallel_2^2,
\end{equation}
where $\{\mathbf{S}_{j}^{*}, \mathbf{U}_{j}^{*}\}$ is one training sample consisting of ground truth landmarks $\mathbf{U}_{j}^{*}$ on the images of a subject and the subject's ground truth frontal and neutral 3D face shape $\mathbf{S}_{j}^{*}$.

A comprehensive description of the SCU-BRL system can be found in the paper Tian et al.~\cite{tian_fg2018_workshop}.

\subsection{3DDFA and 3DMM-CNN}
All the above three 3D face fitting algorithms are 2D facial landmark based.
Although our ESO fitting algorithm exploits texture information for the texture model and lighting model fitting, it does not require texture for shape fitting.
Therefore, it would be interesting to explore more complicated 3D face model fitting algorithms exploiting textural information.
To this end, we evaluate two deep learning based 3D face fitting algorithms, 3DDFA~\cite{DBLP:conf/cvpr/ZhuLLSL16} and 3DMM-CNN~\cite{DBLP:conf/cvpr/TranHMM17}, on the Stirling benchmark.
To be more specific, for 3DDFA, we first use the MTCNN and CNN6 to obtain the 68 facial landmark for an input 2D image.
Then the 68 landmarks are used to initialise the 3DDFA fitting algorithm provided by its authors\footnote{http://www.cbsr.ia.ac.cn/users/xiangyuzhu/projects/3DDFA/main.htm}. As face model, the 3DDFA fitting uses a modified Basel Face Model~\cite{DBLP:conf/avss/PaysanKARV09} with expressions.
We use the term `MTCNN-CNN6-3DDFA' for this system.
For 3DMM-CNN, we use the software provided by the authors\footnote{https://github.com/anhttran/3dmm\_cnn}.
The software adopts the Holistically Constrained Local Model (HCLM) for 2D facial landmark detection~\cite{BMVC2016_95} and the Basel 3D face model for 3D face reconstruction.
We use the term `3DMM-CNN' for the system.

\section{Evaluation results} 
\label{sec:competition_results}
We first compare the average RMSE of different systems on the benchmark dataset.
The results are reported in Table~\ref{table_comparsion}.
\begin{table}[!t]
\centering
\caption{A comparison of different systems in terms of 3D-RMSE.}
\label{table_comparsion}
\begin{tabular}{lccc}
\hline
 Method & HQ & LQ & Full \\
\hline
MTCNN-CNN6-eos & 2.70$\pm$0.88 & 2.78$\pm$0.95 & 2.75$\pm$0.93 \\
MTCNN-CNN6-ESO & 2.68$\pm$0.78 & 2.74$\pm$0.89 & 2.72$\pm$0.86 \\
MTCNN-CNN6-3DDFA & \textbf{2.04$\pm$0.67}  & \textbf{2.19$\pm$0.70} & \textbf{2.14$\pm$0.69} \\
3DMM-CNN & 2.20$\pm$0.61 & 2.38$\pm$0.72 & 2.32$\pm$0.69 \\
SCU-BRL & 2.65$\pm$0.67 & 2.87$\pm$0.81 & 2.81$\pm$0.80 \\
 \hline
\end{tabular}
\end{table}

\begin{figure}[!t]
\centering
\subfloat[High quality set]{
 \label{fig_hq_ced}
 \includegraphics[clip,trim=30mm 100mm 35mm 105mm, width=1\linewidth]{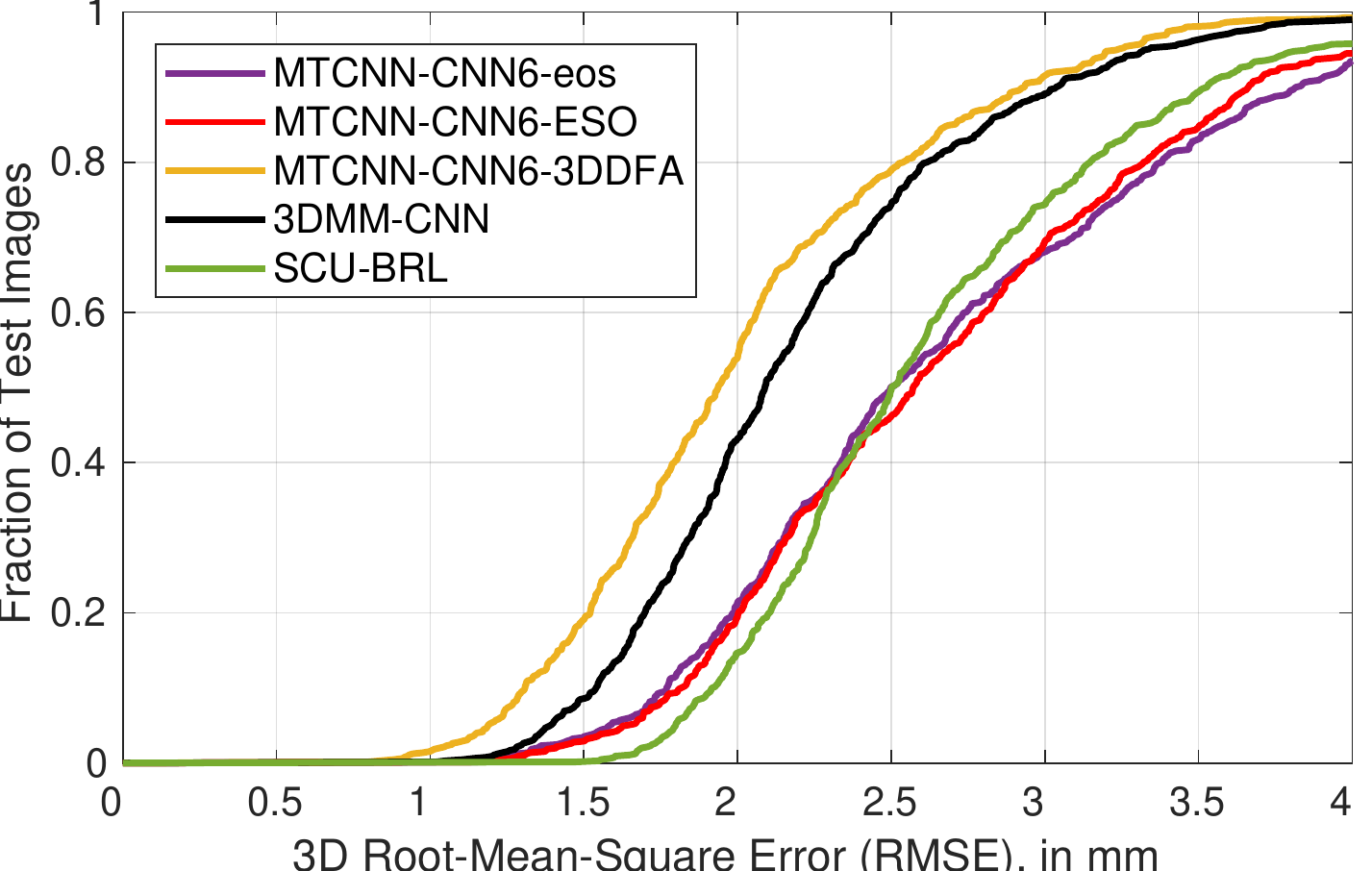}
}

\subfloat[Low quality set]{
 \label{fig_lq_ced}
 \includegraphics[clip,trim=30mm 100mm 35mm 105mm, width=1\linewidth]{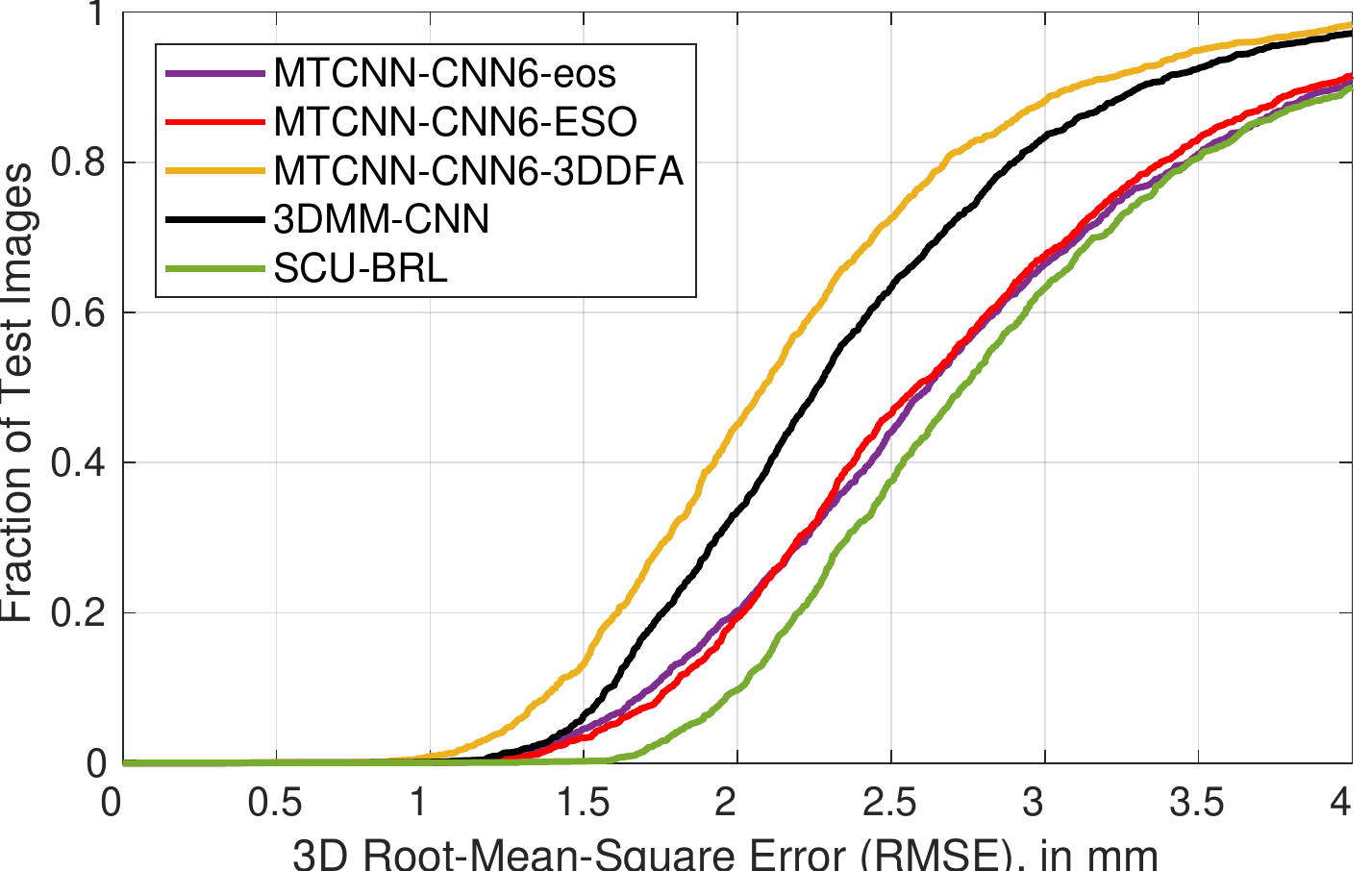}
}

\subfloat[Full set]{
 \label{fig_all_ced}
 \includegraphics[clip,trim=30mm 100mm 35mm 105mm, width=1\linewidth]{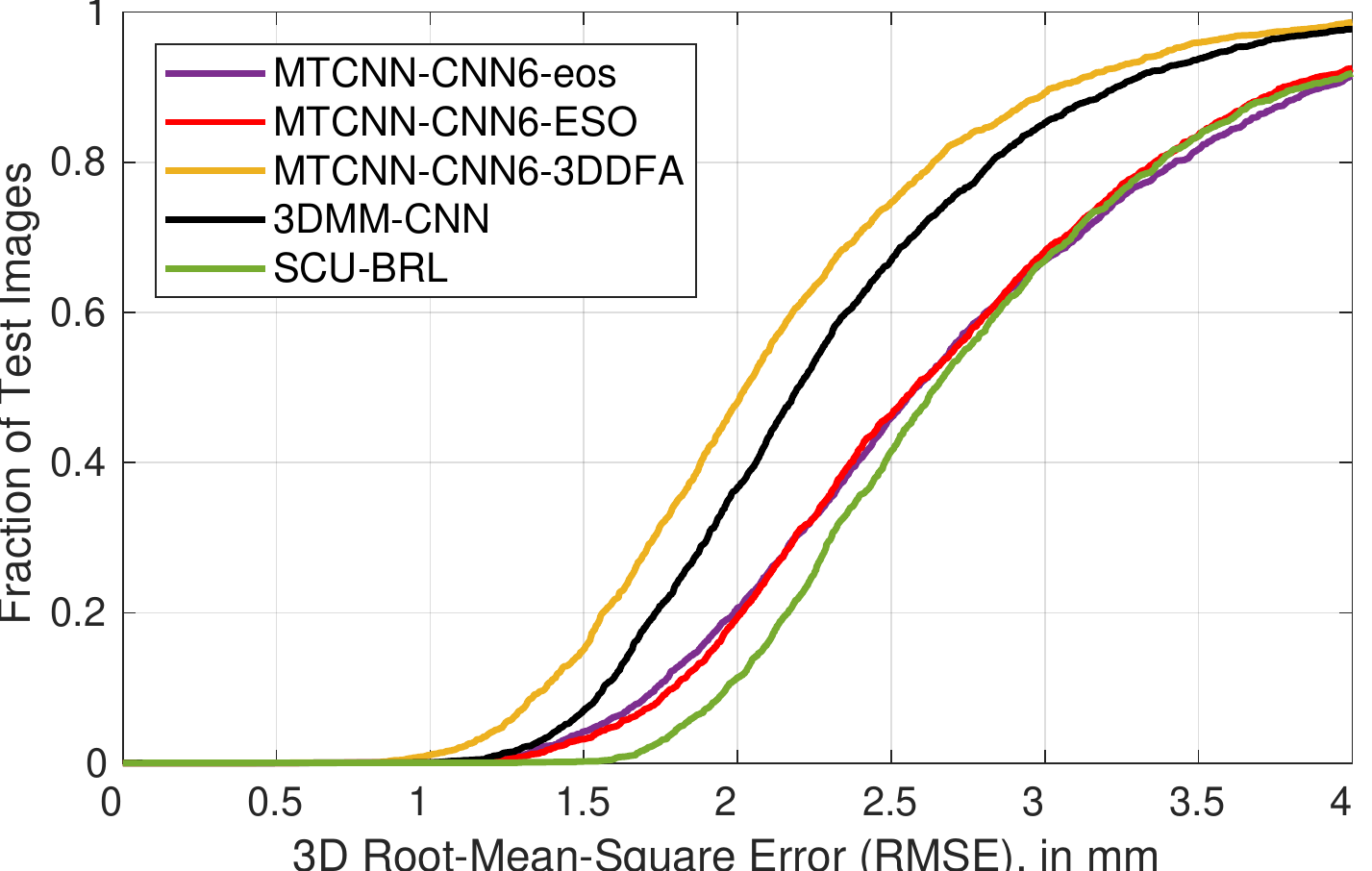}
}
\caption{A comparison of different systems in their CED curves. The results were evaluated on the (a) high quality subset, (b) low quality subset, and (c) full set of the test set selected from the Stirling ESRC dataset.
The evaluation metric is RMSE in millimetre.}
\label{fig_ceds}
\end{figure}

It is clear that the low quality subset is more challenging than the high quality subset.
All the five methods have higher reconstruction errors on the low quality subset.
However, this difference is minor for all the five systems.
The reason is two-fold.
\textit{First}, the high quality subset contains many 2D face images with extreme pose variations, up to $\pm90^\circ$, as well as strong lighting changes thus makes the task more challenging.
\textit{Second}, MTCNN+CNN6+eos, MTCNN+CNN6+ESO and SCU-BRL systems are landmark-only 3D face reconstruction methods and their performance only relies on the accuracy of the detected 2D facial landmarks.
The area of 2D facial landmark detection has already been very well developed for unconstrained facial images in the presence of a variety of image degradation types.
Thus the results of these three landmark-only fitting algorithms measured on the high quality and low quality subsets do not have very big difference.
For the MTCNN+CNN6+3DDFA system, it uses the same state-of-the-art CNN-based face detector and landmark detector, but as initialisation of the 3D face fitting stage of 3DDFA, which is also CNN-based and trained on a large number of unconstrained faces.
In addition, 3DDFA has multiple iterations cascaded for 3D face reconstruction using textural information.
For the 3DMM-CNN, it uses ResNet with 101 layers trained on a large number of real 2D face images for 3D face reconstruction.
The input for 3DMM-CNN is a 2D face image and the outputs are 3D face model parameters.
Both 3DDFA and 3DMM-CNN are deep learning based and trained from a large amount of real face image data hence they can cope with the low quality set very well.
It should be also noted that, in this scenario, the 3DDFA algorithm benefits immensely from the detected 2D facial landmarks by the state-of-the-art CNN6 model.

It should be noted that the SCU-BRL group also conducted multi-image fitting that used all the input 2D image of a subject for 3D face reconstruction.
On average, their system reduces the 3D-RMSE from $2.81\pm0.80$ to $2.26\pm0.72$ on the benchmark dataset by fitting all the input 2D images together for a subject.
This result shows that the SCU-BRL system can effectively utilise the complementary information in multiple images for 3D face reconstruction.
For more details of their multi-image fitting method, please refer to~\cite{tian_fg2018_workshop}.

To better understand the performance of different systems, we also plot the Cumulative Error Distribution (CED) curves of the five systems in Fig.~\ref{fig_ceds}.
The MTCNN+CNN6+3DDFA and 3DMM-CNN systems outperform all the other three systems that only fit a 3D shape model to 2D facial landmarks.
This is an interesting result for the community.
It means that the textural information plays a very important role for high-performance 3D face reconstruction.
However, it should be noted that the fitting of 3DDFA involves multiple iterations with cascaded CNN networks and 3DMM-CNN relies on the very deep ResNet-101 architecture, hence the speed of them cannot satisfy the requirement of a real-time application.
The speed of the 3DDFA implementation provided by its authors is around 1.5 fps tested on a Intel Core i7 6700HQ CPU @ 3.2GHz.
In contrast, the speed of eos is more than 100 fps, which is orders of magnitude faster than 3DDFA.


\section{Conclusion}
A new benchmark dataset was presented in this paper, used for the evaluation of 3D face reconstruction from single 2D face images in the wild.
To this end, a subset of the Stirling ESRC 3D face dataset has been used to create the test set.
The competition was conducted on the real 2D face images of 135 subjects and the evaluation was performed based on their real 3D face ground truth scans.
To facilitate the competition, an evaluation protocol as well as a Python script were provided.

We have compared five state-of-the-art 3D face reconstruction systems on the proposed benchmark dataset, including a system submitted by the Sichuan University, two approaches implemented by the organisers from the University of Surrey and two state-of-the-art deep learning based approaches.
From the performance difference between purely landmark based and texture based reconstruction methods one main conclusion is that texture bolsters a significant amount of extra information about 3D shape. The exploitation of this, however, comes at a price of increased computational time.

The presented benchmark with evaluation data and protocol, together with a comprehensive analysis of different competing algorithms, support future evaluation in the community.

\section*{Acknowledgement}
The authors gratefully acknowledge the support from the EPSRC programme grant (EP/N007743/1), the National Natural Science Foundation of China (61373055, 61672265) and the NVIDIA GPU grant program.
We would also like to convey our great thanks to Dr. Muhammad Awais, Dr. Chi-Ho Chan and Mr. Michael Danner from the University of Surrey, Mr. Hefeng Yin and Mr. Yu Yang from the Jiangnan University for their help on creating the benchmark dataset and revising the paper.



\bibliographystyle{IEEEtran}
\bibliography{references}

\begin{thebibliography}{10}
\providecommand{\url}[1]{#1}
\csname url@samestyle\endcsname
\providecommand{\newblock}{\relax}
\providecommand{\bibinfo}[2]{#2}
\providecommand{\BIBentrySTDinterwordspacing}{\spaceskip=0pt\relax}
\providecommand{\BIBentryALTinterwordstretchfactor}{4}
\providecommand{\BIBentryALTinterwordspacing}{\spaceskip=\fontdimen2\font plus
\BIBentryALTinterwordstretchfactor\fontdimen3\font minus
  \fontdimen4\font\relax}
\providecommand{\BIBforeignlanguage}[2]{{%
\expandafter\ifx\csname l@#1\endcsname\relax
\typeout{** WARNING: IEEEtran.bst: No hyphenation pattern has been}%
\typeout{** loaded for the language `#1'. Using the pattern for}%
\typeout{** the default language instead.}%
\else
\language=\csname l@#1\endcsname
\fi
#2}}
\providecommand{\BIBdecl}{\relax}
\BIBdecl

\bibitem{DBLP:conf/siggraph/BlanzV99}
V.~Blanz and T.~Vetter, ``{A Morphable Model for the Synthesis of 3D Faces},''
  in \emph{the 26th Annual Conference on Computer Graphics and Interactive
  Techniques (SIGGRAPH)}, W.~N. Waggenspack, Ed., 1999, pp. 187--194.

\bibitem{feng2015cascaded}
Z.-H. Feng, G.~Hu, J.~Kittler, W.~Christmas, and X.-J. Wu, ``Cascaded
  collaborative regression for robust facial landmark detection trained using a
  mixture of synthetic and real images with dynamic weighting,'' \emph{IEEE
  Transactions on Image Processing}, vol.~24, no.~11, pp. 3425--3440, 2015.

\bibitem{kittler20163d}
J.~Kittler, P.~Huber, Z.-H. Feng, G.~Hu, and W.~Christmas, ``3d morphable face
  models and their applications,'' in \emph{9th International Conference on
  Articulated Motion and Deformable Objects (AMDO)}, vol. 9756, 2016, pp.
  185--206.

\bibitem{zeng2017examplar}
D.~Zeng, Q.~Zhao, S.~Long, and J.~Li, ``Examplar coherent 3d face
  reconstruction from forensic mugshot database,'' \emph{Image and Vision
  Computing}, vol.~58, pp. 193--203, 2017.

\bibitem{DBLP:conf/icb/LiuHSWZ17}
F.~Liu, J.~Hu, J.~Sun, Y.~Wang, and Q.~Zhao, ``Multi-dim: {A} multi-dimensional
  face database towards the application of 3d technology in real-world
  scenarios,'' in \emph{2017 {IEEE} International Joint Conference on
  Biometrics (IJCB)}, 2017, pp. 342--351.

\bibitem{koppen2018gaussian}
P.~Koppen, Z.-H. Feng, J.~Kittler, M.~Awais, W.~Christmas, X.-J. Wu, and H.-F.
  Yin, ``Gaussian mixture 3d morphable face model,'' \emph{Pattern
  Recognition}, vol.~74, pp. 617--628, 2018.

\bibitem{DBLP:conf/cvpr/ZhuLLSL16}
X.~Zhu, Z.~Lei, X.~Liu, H.~Shi, and S.~Z. Li, ``{Face Alignment Across Large
  Poses: {A} 3D Solution},'' in \emph{IEEE Conference on Computer Vision and
  Pattern Recognition CVPR}, 2016, pp. 146--155.

\bibitem{DBLP:conf/iccv/BulatT17}
A.~Bulat and G.~Tzimiropoulos, ``{How Far are We from Solving the 2D {\&} 3D
  Face Alignment Problem? (and a Dataset of 230, 000 3D Facial Landmarks)},''
  in \emph{{IEEE} International Conference on Computer Vision (ICCV)}, 2017,
  pp. 1021--1030.

\bibitem{liu2017dense}
Y.~Liu, A.~Jourabloo, W.~Ren, and X.~Liu, ``Dense face alignment,'' \emph{arXiv
  preprint arXiv:1709.01442}, 2017.

\bibitem{Bagdanov:2011:FHF:2072572.2072597}
A.~D. Bagdanov, A.~Del~Bimbo, and I.~Masi, ``{The Florence 2D/3D Hybrid Face
  Dataset},'' in \emph{the Joint ACM Workshop on Human Gesture and Behavior
  Understanding}, 2011, p. 79–80.

\bibitem{DBLP:journals/cg/HernandezHCM17}
M.~Hernandez, T.~Hassner, J.~Choi, and G.~G. Medioni, ``{Accurate 3D face
  reconstruction via prior constrained structure from motion},''
  \emph{Computers {\&} Graphics}, vol.~66, pp. 14--22, 2017.

\bibitem{DBLP:conf/cvpr/TranHMM17}
A.~T. Tran, T.~Hassner, I.~Masi, and G.~G. Medioni, ``{Regressing Robust and
  Discriminative 3D Morphable Models with a Very Deep Neural Network},'' in
  \emph{IEEE Conference on Computer Vision and Pattern Recognition (CVPR)},
  2017, pp. 1493--1502.

\bibitem{DBLP:journals/corr/JacksonBAT17}
A.~S. Jackson, A.~Bulat, V.~Argyriou, and G.~Tzimiropoulos, ``{Large Pose 3D
  Face Reconstruction from a Single Image via Direct Volumetric {CNN}
  Regression},'' \emph{arXiv}, 2017.

\bibitem{DBLP:conf/eccv/JeniTYSC16}
L.~A. Jeni, S.~Tulyakov, L.~Yin, N.~Sebe, and J.~F. Cohn, ``{The First 3D Face
  Alignment in the Wild {(3DFAW)} Challenge},'' in \emph{Europearn Conference
  on Computer Vision Workshops - (ECCVW)}, G.~Hua and H.~J{\'{e}}gou, Eds.,
  vol. 9914, 2016, pp. 511--520.

\bibitem{DBLP:journals/corr/BoothAPTPZ17}
J.~Booth, E.~Antonakos, S.~Ploumpis, G.~Trigeorgis, Y.~Panagakis, and
  S.~Zafeiriou, ``{3D Face Morphable Models "In-the-Wild"},'' \emph{arXiv},
  2017.

\bibitem{DBLP:conf/visapp/HuberHTMKCRK16}
P.~Huber, G.~Hu, J.~R. Tena, P.~Mortazavian, W.~P. Koppen, W.~J. Christmas,
  M.~R{\"{a}}tsch, and J.~Kittler, ``{A Multiresolution 3D Morphable Face Model
  and Fitting Framework},'' in \emph{the 11th Joint Conference on Computer
  Vision, Imaging and Computer Graphics Theory and Applications (VISIGRAPP)},
  2016, pp. 79--86.

\bibitem{tian_fg2018_workshop}
W.~Tian, F.~Liu, and Q.~Zhao, ``{Landmark-based 3D Face Reconstruction from an
  Arbitrary Number of Unconstrained Images},'' in \emph{IEEE International
  Conference and Workshops on Automatic Face and Gesture Recognition (FG)},
  2018.

\bibitem{7553523}
K.~Zhang, Z.~Zhang, Z.~Li, and Y.~Qiao, ``Joint face detection and alignment
  using multitask cascaded convolutional networks,'' \emph{IEEE Signal
  Processing Letters}, vol.~23, no.~10, pp. 1499--1503, 2016.

\bibitem{feng2017face}
Z.-H. Feng, J.~Kittler, M.~Awais, P.~Huber, and X.-J. Wu, ``{Face Detection,
  Bounding Box Aggregation and Pose Estimation for Robust Facial Landmark
  Localisation in the Wild},'' in \emph{IEEE Conference on Computer Vision and
  Pattern Recognition Workshops (CVPRW)}, 2017, pp. 160--169.

\bibitem{feng2017dynamic}
Z.-H. Feng, J.~Kittler, W.~Christmas, P.~Huber, and X.-J. Wu, ``Dynamic
  attention-controlled cascaded shape regression exploiting training data
  augmentation and fuzzy-set sample weighting,'' in \emph{The IEEE Conference
  on Computer Vision and Pattern Recognition (CVPR)}, 2017, pp. 2481--2490.

\bibitem{feng2018wing}
Z.-H. Feng, J.~Kittler, M.~Awais, P.~Huber, and X.-J. Wu, ``Wing loss for
  robust facial landmark localisation with convolutional neural networks,'' in
  \emph{IEEE Conference on Computer Vision and Pattern Recognition (CVPR)},
  2018.

\bibitem{le2012interactive}
V.~Le, J.~Brandt, Z.~Lin, L.~Bourdev, and T.~S. Huang, ``Interactive facial
  feature localization,'' in \emph{European Conference on Computer Vision
  (ECCV)}.\hskip 1em plus 0.5em minus 0.4em\relax Springer, 2012, pp. 679--692.

\bibitem{belhumeur2013localizing}
P.~N. Belhumeur, D.~W. Jacobs, D.~J. Kriegman, and N.~Kumar, ``Localizing parts
  of faces using a consensus of exemplars,'' \emph{IEEE Transactions on Pattern
  Analysis and Machine Intelligence}, vol.~35, no.~12, pp. 2930--2940, 2013.

\bibitem{sagonas2013300}
C.~Sagonas, G.~Tzimiropoulos, S.~Zafeiriou, and M.~Pantic, ``300 faces
  in-the-wild challenge: The first facial landmark localization challenge,'' in
  \emph{IEEE International Conference on Computer Vision Workshops (ICCVW)},
  2013, pp. 397--403.

\bibitem{zhu2012face}
X.~Zhu and D.~Ramanan, ``Face detection, pose estimation, and landmark
  localization in the wild,'' in \emph{IEEE Conference on Computer Vision and
  Pattern Recognition (CVPR)}, 2012, pp. 2879--2886.

\bibitem{gross2010multi}
R.~Gross, I.~Matthews, J.~Cohn, T.~Kanade, and S.~Baker, ``Multi-pie,''
  \emph{Image and Vision Computing}, vol.~28, no.~5, pp. 807--813, 2010.

\bibitem{feng2016unified}
Z.-H. Feng, J.~Kittler, W.~Christmas, and X.-J. Wu, ``A unified tensor-based
  active appearance face model,'' \emph{arXiv}, 2016.

\bibitem{hu2017efficient}
G.~Hu, F.~Yan, J.~Kittler, W.~Christmas, C.-H. Chan, Z.-H. Feng, and P.~Huber,
  ``Efficient 3d morphable face model fitting,'' \emph{Pattern Recognition},
  vol.~67, pp. 366--379, 2017.

\bibitem{DBLP:conf/avss/PaysanKARV09}
P.~Paysan, R.~Knothe, B.~Amberg, S.~Romdhani, and T.~Vetter, ``A 3d face model
  for pose and illumination invariant face recognition,'' in \emph{IEEE
  International Conference on Advanced Video and Signal Based Surveillance
  {AVSS}}, 2009, pp. 296--301.

\bibitem{BMVC2016_95}
A.~Z. L.-P.~M. Kanggeon~Kim, Tadas~Baltru\v{s}aitis and G.~Medioni,
  ``Holistically constrained local model: Going beyond frontal poses for facial
  landmark detection,'' in \emph{Proceedings of the British Machine Vision
  Conference (BMVC)}, E.~R.~H. Richard C.~Wilson and W.~A.~P. Smith, Eds.\hskip
  1em plus 0.5em minus 0.4em\relax BMVA Press, September 2016, pp. 95.1--95.12.

\end{thebibliography}
%



\end{document}